# Machine Learning:
## When and Where the Horses Went Astray?

Emanuel Diamant
*VIDIA-mant, Israel*
emanl.245@gmail.com
http://www.vidia-mant.info

## 1. Introduction

The year of 2006 was exceptionally cruel to me – almost all of my papers submitted for that year conferences have been rejected. Not "just rejected" – unduly strong rejected. Reviewers of the ECCV (European Conference on Computer Vision) have been especially harsh: "This is a philosophical paper... However, ECCV neither has the tradition nor the forum to present such papers. Sorry..." O my Lord, how such an injustice can be tolerated in this world? However, on the other hand, it can be easily understood why those people hold their grudges against me: Yes, indeed, I always try to take a philosophical stand in all my doings: in thinking, paper writing, problem solving, and so no. In my view, philosophy is not a swear-word. Philosophy is a keen attempt to approach the problem from a more general standpoint, to see the problem from a wider perspective, and to yield, in such a way, a better comprehansion of the problem's specificity and its interaction with other world realities. Otherwise we are doomed to plunge into the chasm of modern alchemy – to sink in partial, task-oriented determinations and restricted solution-space explorations prone to dead-ends and local traps.

It is for this reason that when I started to write about "Machine Learning", I first went to the Wikipedia to inquire: What is the best definition of the subject? "Machine Learning is a subfield of Artificial Intelligence" – was the Wikipedia's prompt answer. Okay. If so, then: "What is Artificial Intelligence?" – "Artificial Intelligence is the intelligence of machines and the branch of computer science which aims to create it" – was the response. Very well. Now, the next natural question is: "What is Machine Intelligence?" At this point, the kindness of Wikipedia has been exhausted and I was thrown back, again to the Artificial Intelligence definition. It was embarrassing how quickly my quest had entered into a loop – a little bit confusing situation for a stubborn philosopher.

Attempts to capitalize on other trustworthy sources were not much more productive (Wang, 2006; Legg & Hutter, 2007). For example, Hutter in his manuscript (Legg & Hutter, 2007) provides a list of 70-odd "Machine Intelligence" definitons. There is no consensus among the items on the list – everyone (and the citations were chosen from the works of the most prominent scholars currently active in the field), everyone has his own particular view on the subject. Such inconsistency and multiplicity of definitions is an unmistakable sign of philosophical immaturity and a lack of a will to keep the needed grade of universality and generalization.

It goes without saying, that the stumbling-block of the Hutter's list of definitions (Legg & Hutter, 2007) are not the adjectives that was used – after all the terms "Artificial" and "Machine" are consensually close in their meaning and therefore are commonly used interchangeably. The real problem – is the elusive and indefinable term „Intelligence".

I will not try the readers' patience and will not tediously explain how and why I had arrived at my own definition of the matters that I intend to scrutinize in this paper.

I hope that my philosophical leanings will be generously excused and the benevolent readers will kindly accept the unusual (reverse) layout of the paper's topics. For the reasons that would be explained in a little while, the main and the most general paper's idea will be presented first while its compiling details and components will be exposed (in a discending order) afterwards. And that is how the proposed paper's layout should look like:

- **Intelligence is the system's ability to process information**. This statement is true both for all biological natural systems as for artificial, human-made systems. By "information processing" we do not mean its simplest forms like information storage and retrieval, information exchange and communication. What we have in mind are the high-level information processing abilities like information analysis and interpretation, structure patterns recognition and the system's capacity to make decisions and to plan its own behavior.
- **Information in this case should be defined as a description** – A language and/or an alphabet-based description, which results in a reliable reconstruction of an original object (or an event) when such a description is carried out, like an execution of a computer program.
- **Generally, two kinds of information must be distinguished:** Objective (physical) information and subjective (semantic) information. By physical information we mean the description of data structures that are discernable in a data set. By semantic information we mean the description of the relationships that may exist between the physical structures of a given data set.
- **Machine Learning** is defined as the best means for appropriate information retrieval. Its usage is endorsed by the following fundamental assumptions: 1) Structures can be revealed by their characteristic features, 2) Feature aggregation and generalization can be achieved in a bottom-up manner where final results are compiled from the component details, 3) Rules, guiding the process of such compilation, could be learned from the data itself.
- **All these assumptions validating Machine Learning applications are false.** (Further elaboration of the theme will be given later in the text). Meanwhile the following considerations may suffice:
- Physical information, being a natural property of the data, can be extracted instantly from the data, and any special rules for such task accomplishment are not needed. Therefore, Machine Learning techniques are irrelevant for the purposes of physical information retrieval.
- Unlike physical information, semantics is not a property of the data. Semantics is a property of an external observer that watches and scrutinizes the data. Semantics is assigned to phisical data structures, and therefore it can not be learned straightforwardly from the data. For this reason, Machine Learning techniques are useless and not applicable for the purposes of smantic information extraction. Semantics is a shared convention, a mutual agreement between the members of a particular group of viewers or users. Its assignment has to be done on the basis of a consensus knowledge that is shared among the group members, and which an artificial semantic-processing system has to possess at its disposal. Accomodation and fitting of this knowledge presumes availability of a different and usually overlooked special learning technique, which would be best defined as **Machine Teaching –** a technique that would facilitate externally-prepared-knowledge transfer to the system's disposal .

These are the topics that I am interested to discuss in this paper. Obviously, the reverse order proposed above, will never be reified – there are paper organization rules and requirements, which none never will be allowed to override. They must be, thus, reverently obeyed. And I earnestly promiss to do this (or at least to try to do this) in this paper.

## 2. When the State of the Art is Irrelevant

One of the commonly accepted rules prescribes that the Introduction Section has to be succeeded by a clear presentation of a following subject: What is the State of the Art in the field and what is the related work done by the other researchers? Unfortunately, I'm unable to meet this requirement, because (to the best of my knowledge) there is no relevant work in the field that can be used for this purpose. Or, let us put this in a more polite way: The work presented in this paper is so different from other mainstream approaches that it would be unwise to compare it with the rest of the work in the field and to discuss arguments in favour or against their endless disagreements and discrepancies. However, to avoid any possible allegations in disrespectfulness, I would like to provide here some reflections on the departure points of my work, which (I hope) are common to many friends and foes in the domain.
My first steps in the field were inspired by David Marr's ideas about the "Primal" and the "Two-and-a-half" image representation sketch, which is carrying out the information content of an image (Marr, 1978; Marr, 1982). Image understanding was always the most relevant and the most palpable manifestation of human intelligence, and so, those who are busy with intelligence replications in machines, are due to cope with image understanding and image processing issues.
"You see, – had I proudly agitated my employers, trying to convince them to fund my image-processing enterprises, – meagre lines of a painter's caricature provide you with all information needed to comprehend the painter's intention and to easily recognise the objects drawn in the picture. Edges are the information bearers! Edge exploration and processing will help us to reach advances in pattern recognition and image understanding. "
My employers were skeptic and penny-pinching, but nevertheless, I was allowed to do some work. However, very soon it had become clear that my problems are far from being information retrieval issues – my real problem was to run (approximately in a real-time fashion) a 3-by-3 (or 5-by-5) operator over a 256-by-256 pixel image. And only then, when the run is somehow successfully completed, I was doomed to find myself inflated with a multitude of edges: cracked, disjoint, and inconsistent. On one hand, an entire spectrum of dissimilar edge pieces, and on the other hand – a striking deficit of any hints regarding how to arrange them into something handy and meaningful. At least, to choose among them (to discriminate, to segment, to threshold) those that would be suitable for further processing. Even though, it was at all not sure that anybody knows what such a processing should be.

It was not only my nightmare. Many people have swamped in this bog. Many are still trying to tempt the fate – even today, the flow of edge extraction and segmentation publications does not dry up, and new machine learning techniques are reportedly proposed to cure the problem (Ghosh et al., 2007; Awad & Man, 2008; Qiu & Sun, 2009).

Human vision physiology studies, which have been always seen as an endless source of computer vision R&D inspiration, have also proved to be of a little help here. Treisman's feature-integration theory (Treisman & Gelade, 1980) and Biederman's recognition-by-components theory (Biederman, 1987) – the cornerstones of contemporary vision science – were fitting well the bottom-up image processing philosophy, (where initial feature gathering is followed by further feature consolidation), but they have nothing to say about how this feature aggregation and integration (into meaningful perceptible objects) has to be realized. They only say that this process has to be done in a top-down fashion, in opposite to the bottom-up processing of the initial features.

To overcome the problem, a great variety of so-called "binding" theories have been proposed (Treisman, 1996; Treisman, 2003). However, all of them turned out as inappropriate. In a desperate attempt to resolve this irresolvable contradiction, even a theory of a mysterious homunculus has been proposed – a "little man inside the head" that perceives the world through our senses and then unmistakably fulfils all the needed (intelligent) actions (Crick & Koch, 2000). But the theory of the homunculus has not taken roots. Human level intelligence has been and continues to be a challenge, and nothing in the field has changed since the 50s of the past century, when the first steps of Artificial Intelligence exploration have been carried out (Turing, 1950; McCarthy et al., 1955).

## 3. In Search for a Better Fortune

I am not trying to claim that I am more clever or wise than others. All the stupid things that others have persistently tried to do, I have repeatedly tried as well. But in one thing, however, I was certainly different from the others – I have never neglected my final goal: To grasp the information content of an image. Together with other image-processing "partisans" and "camarados" I fought my pixel-oriented battles, but a dream about object-oriented image processing was always blooming in my heart.

As you can understand, nothing worthy had come out from that. Nevertheless, some of the things that I was lucky to make happen (at that time) are worth to be mentioned here. For example, I have invented a notion of "Single Pixel Information Content" and a way for its quantitative evaluation (Diamant, 2003). I have also invented a notion of "Specific Information Density of an Image", and, relying on the pixel's information content (measure), I have attempted to investigate the effect of "Image Information Content Conservation". That is, when an image scale is successively reduced, Image Specific Information Density remains unchanged (or even slightly grows up). Then, at some level of reduction, it rapidly declines. This scale, actually the scale one step preceding the drop of Information Density, I thought, should be the most advantageous (scale) to start image information content explorations.

A paper, containing quantitative results and a proof of this idea, has been submitted to the British Machine Vision Conference (Diamant, 2002), but, (as usually), was decisively rejected. Never mind, these investigations have led to an important insight that image information content excavation has to be commenced at an optimal, low-dimensional image representation scale.

I am proud to inform the interested readers that similar investigations have been performed recently (and similar results have been attained) by MIT researchers (Torralba, 2009). However, that was done about seven years later, and only in qualitative experiments conducted on human participants (but not as a quantitative work).

Never mind, the idea of initial low-dimensional image exploration was in some way consistent with a naïve psychological vision conjecture about how humans look at a scene. Since biological vision research was always busy with only foveated vision studies, one principal aspect of human vision was always remained neglected: How does the brain know where to look in a scene? We do not search our field of view in a regular, raster-scan manner. On the contrary, we do this in an unpredictable, but certainly a not-random manner (Koch et al., 2007; Shomstein & Behrmann, 2008). If so, how does the brain know where to place the eye's fovea – (the main means for visual information gathering) – before it knows in advance where such information is to be found? Certainly, the brain must have a prior knowledge about the scene layout, about the general map of a scene. Certainly, the scale of this map must be several orders lower than the fovea resolution scale, and it is clear that these information gathering maps are being used simultaneously and interchangeably.

Such considerations have inevitably led us to a conclusion that other theories, currently unknown to us, which would be capable of explaining such multiscale brain performance have to be urgently searched for. Indeed, very soon I came upon an appropriate theory. And even not a single one, but a whole bundle of theories.

In the middle of the 60s of the previous century, three almost simultaneous, but absolutely independently developed, theories have sprung up: Solomonoff's theory of Inference (Solomonoff, 1964), Kolmogorov's Complexity theory (Kolmogorov, 1965), and Chaitin's Algorithmic Information theory (Chaitin, 1966). Since among the three, Kolmogorov's theory is the most known one, I will first and mainly refer to it in our further discussion.

Just as Shannon's Information theory (Shannon, 1948) published almost 20 years ahead, Kolmogorov's theory was aimed at providing means for measuring "information". However, while Shannon's theory was dealing only with the average amount of information conveyed by an outcome of a random source, Kolmogorov's theory was busy with information contained in a particular isolated object. Thus, Kolmogorov's theory was far more suitable to deal with human vision peculiarities.

However, I do not intend to bother the readers with explanations about Kolmogorov's theory merits. Such expanded enlightenment could be found else where, for example (Li & Vitanyi, 2008; Grunvald & Vitanyi, 2008). My humble intention is, relying on the insights of the Kolmogorov's theory, to provide some useful illuminations, which can be deduced from the theory and applied to the practice of image information content excavation.

An essential part of my work has been already done in the past years, and has been even published on several occasions (Diamant, 2004; Diamant, 2005a; Diamant, 2005b). (The publications could be easily found at some freely accessible web repositories, like CiteSeer, Eprintweb, ArXiv, etc. And also on my personal web site: http://www.vidia-mant.info).

However, for the consistency of our discussion, I would like to repeat here the main points of these previous publications.

The key point is that **information is a description**, a certain alphabet-based or language-based description, which Kolmogorov's theory regards as a program that, being executed, trustworthy reproduces the original object (Vitanyi, 2006). In an image, such objects are visible data structures from which an image is comprised of. So, a set of reproducible descriptions of image data structures is the information contained in an image.

The Kolmogorov's theory prescribes the way in which such descriptions must be created: At first, the most simplified and generalized structure must be described. Recall the Occam's Razor principle: Among all hypotheses consistent with the observation choose the simplest one that is cohirent with the data, (Sadrzadeh, 2008). Then, as the level of generalization is gradually decreased, more and more fine-grained image details (structures) become revealed and depicted. This is the second important point, which follows from the theory's pure mathematical considerations: Image **information is a hierarchy of decreasing level descriptions** of information details, which unfolds in a coarse-to-fine top-down manner. (Attention, please! Any bottom-up processing is not mentioned here! There is no low-level feature gathering and no feature binding!!! The only proper way for image information elicitation is a top-down coarse-to-fine way of image processing!)

The third prominent point, which immediately pops-up from the two just mentioned above, is that the top-down manner of image **information elicitation does not require incorporation of any high-level knowledge** for its successful accomplishment. It is totally free from any high-level guiding rules and inspirations. (The homunculus have lost his job and is finally fired).

That is why I call the information, which unconditionally can be found in an image, – the **Physical Information.** That is, information that reflects objective (physical) structures in an image and is totally independent of any high level interpretation of the interrelashions between them.

What immediately follows from this is that high-level image semantics is not an integrated part of image information content (as it is traditionally assumed). It cannot be seen more as a natural property of an image. **Semantic Information**, therefore, must be seen as a property of a human observer that watches and scrutinizes an image. That is why we can say now: **Semantics is assigned to an image by a human observer**. That is strongly at variance with the contemporary views on the concept of semantic information.

As it was mentioned above, I have no intention to argue with the mainstream experts, conference chaires, keynotes speekers and invited talks presenters about the validity of my contemplations, about my philosophical inclinations or research duties and preferences. These respected gentlemans would continue to teach you **how to extract semantic information from an image** or **how it should be derived from low-level information features.**

(I do not provide here examples of such claims. I hope**,** the readers are well enough acquinted with the state of the art in the field and its mainstream developments, to be able to recall the appropriate cases by themselves. I also hope that readers of this paper are ready to change their minds – fifty or so years of Machine Learning triumfal marching in the head of the Artificial Intelligence parade have not got us closer to the desired goal of Intelligent Machines deployment and use. Partially, the loosely defined (or it would be right to say, undefined) departure points of this enterprise were the main reasons responsible for this years-long wandering in the desert far away from the promised land.)

## 4. "Repetitio est Mater Studiorum"

(For those who are not fluent enough in Latin, the translation of this proverb would be: Reiteration is the mother of learning). Okay, I am really sorry that instead of dealing with the declared subject of this paper (that is, Machine Learning and all its corresponding issues), I have to return again and again to topics that have been already discussed in the past and even published at some previous occasions. (But that is the bad luck of an image-processing partisan). Therefore, with all apologies to be due, I will continue our discourse with some extended citations seized from my previously published papers

### 4.1 Image Physical information Processing

The first citation is related to physical information processing issues and is taken from a five years old paper (Diamant, 2004). The citation subject is – an algorithmic implementation of image physical information mining principles.

The algorithm's block-scheme looks as follows:

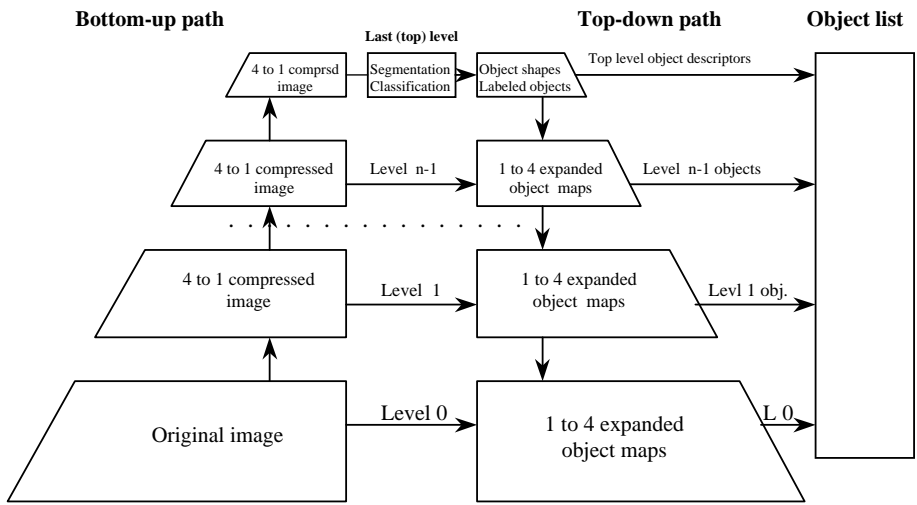

Fig. 1. The block-diagram of physical information elucidation.

As can be seen at Fig. 1, the proposed schema is comprised of three main processing paths: the bottom-up processing path, the top-down processing path and a stack where the discovered information content (the generated descriptions of it) is actually accumulated. The algorithm's structure reflects the principles of information representation, which have been already defined previously.

As it is shown in the schema, the input image is initially squeezed to a small size of approximately 100 pixels. The rules of this shrinking operation are very simple and fast: four non-overlapping neighbor pixels in an image at level $L$ are averaged and the result is assigned to a pixel in a higher ($L$+1)-level image. This is known as "four children to one parent relationship". Then, at the top of the shrinking pyramid, the image is segmented, and each segmented region is labeled. Since the image size at the top is significantly reduced and since in the course of the bottom-up image squeezing a severe data averaging is attained, the image segmentation/labeling procedure does not demand special computational resources. Any well-known segmentation methodology will suffice. We use our own proprietary technique that is based on a low-level (single pixel) information content evaluation (Diamant, 2003), but this is not obligatory.

From this point on, the top-down processing path is commenced. At each level, the two previously defined maps (average region intensity map and the associated label map) are expanded to the size of an image at the nearest lower level. Since the regions at different hierarchical levels do not exhibit significant changes in their characteristic intensity, the majority of newly assigned pixels are determined in a sufficiently correct manner. Only pixels at region borders and seeds of newly emerging regions may significantly deviate from the assigned values. Taking the corresponding current-level image as a reference (the left-side unsegmented image), these pixels can be easily detected and subjected to a refinement cycle. In such a manner, the process is subsequently repeated at all descending levels until the segmentation/classification of the original input image is successfully accomplished.

At every processing level, every image object-region (just recovered or an inherited one) is registered in the objects' appearance list, which is the third constituting part of the proposed scheme. The registered object parameters are the available simplified object's attributes, such as size, center-of-mass position, average object intensity and hierarchical and topological relationship within and between the objects ("sub-part of…", "at the left of…", etc.). They are sparse, general, and yet specific enough to capture the object's characteristic features in a variety of descriptive forms.

Examples of algorithm's performance and some concrete palpable results are provided in several previously published papers (Diamant, 2005a; Diamant, 2005b).

In our current discussion it is worth to be mentioned that: First, image segmentation (physical image structures delineation, physical image information elicitation) is performed in a top-down manner, not in a conventional bottom-up mode. Second, the suggested image segmentation principle does not require any knowledge about high-level rules, which are used to support the segmentation process and which are an obligatory part of any bottom-up proceeding procedure. Third, canceling the necessity of these high-level rules, makes all Machine Learning techniques useless and invalidates all efforts that are specially carried out to meet this sacred requirement! In this way, Machine Learning loses its role as the main performer in physical information recovery enterprises.

**4.2 Image Semantic Information Processing**

The context of this sub-section is also an extended quotation from a recently published paper (Diamant, 2008). The key point of this quotation is a semantic information processing architecture based on the same information-defining rules and the same (top-down) information representation principles that were already introduced in Section 3. The block-schema of such a semantic information processing architecture is borrowed from the above mentioned paper (Diamant, 2008), and is depicted in the Fig. 2.

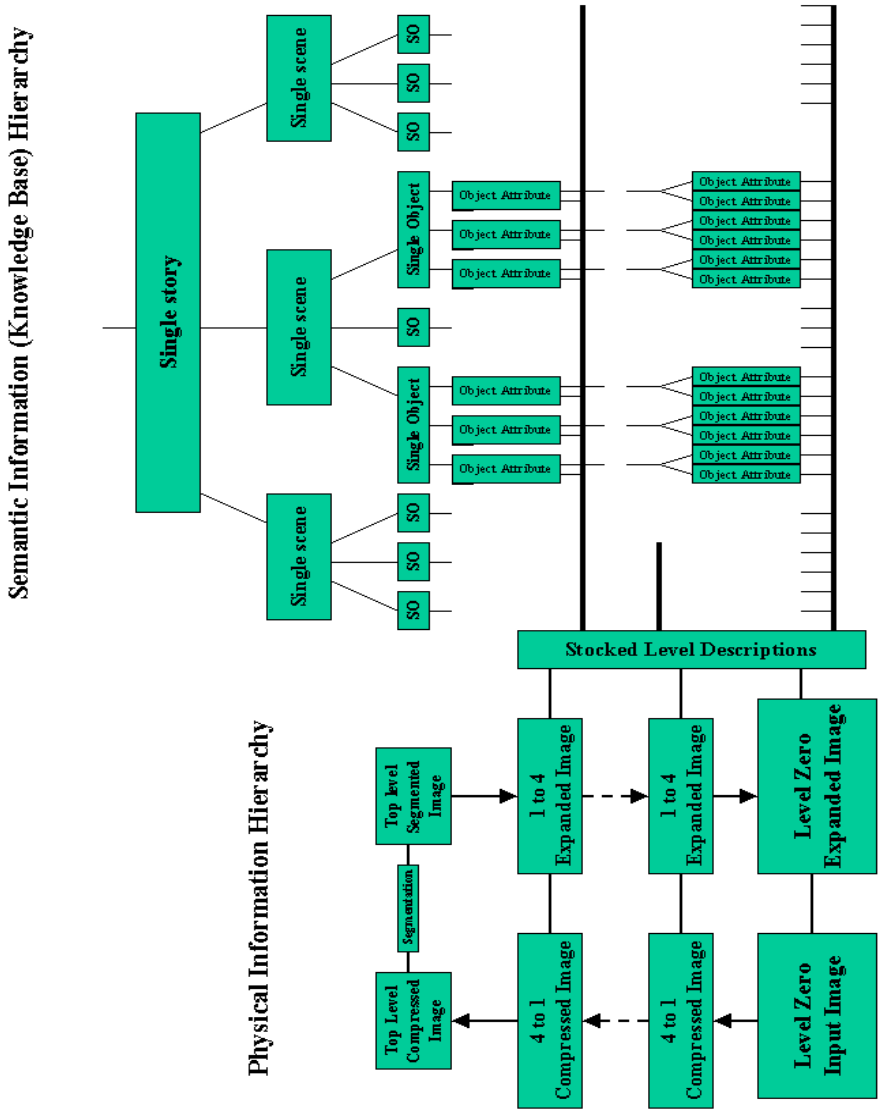

Fig. 2. Physical and Semantic Information processing hierarchies.

Scrutinizing of this general image information processing architecture must be preceded by some remarks: Semantic information, which (as we understand now) is a property of an external observer, is separated and dissociated from the physical information processing, in our case an image. Therefore it must be treated (or modeled) in accordance with observer-specific (his/her) cognitive information processing rules.

It is well known that human cognitive abilities (including the aptness for image interpretation and the capacity to assign semantics to an image) are empowered by the existence of a huge knowledge base about the things in the surrounding world kept in human brain.

This knowledge base is permanently upgraded and updated during the human's life span. So, if we intend to endow our design with some cognitive capabilities we have to provide it with something equivalent to this (human) knowledge base.

It goes without saying that this knowledge base will never be as large and developed as its human prototype. But we are not sure that such a requirement is valid here. After all, humans are also not equal in their cognitive capacities, and the content of their knowledge bases is very diversified as well. (The knowledge base of an aerial photographs interpreter is certainly different from the knowledge base of an X-ray images interpreter, or an IVUS images interpreter, or PET images). The knowledge base of our visual thinking machine has to be small enough to be effective and manageable, but sufficiently large to ensure the machine acceptable performance. Certainly, for our feasibility study we can be satisfied even with a relatively small, specific-task-oriented knowledge base.

The next crucial point is the knowledge representation issue. To deal with it, we first of all must arrive at a common agreement about what is the meaning of the term "knowledge". (A question that usually does not have a single answer.) We state that in our case a suitable definition of it would be: "**Knowledge is memorized information**". Consequently, we can say that knowledge (like information) must be a hierarchy of descriptive items, with the grade of description details growing in a top-down manner at the descending levels of the hierarchy.

One more point that must be mentioned here, is that these descriptions have to be implemented in some alphabet (as it is in the case of physical information) or in a description language (which better fits the semantic information case). Any farther argument being put aside, we will declare that the most suitable language in our case is the natural human language. After all, the real knowledge bases that we are familiar with are implemented in natural human languages.

The next step, then, is predetermined: if natural language is a suitable description implement, the suitable form of this implementation is a narrative, a story tale (Tuffield et al., 2005). If the description hierarchy can be seen as an inverted tree, then the branches of this tree are the stories that encapsulate human's experience with the surrounding world. And the leaves of these branches are single words (single objects) from which the story parts (single scenes) are composed of.

The descent into description details, however, does not stop here, and each single word (single object) can be farther decomposed into its attributes and rules that describe the relations between the attributes.

At this stage the physical information reappears. Because the words are usually associated with physical objects in the real world, words' attributes must be seen as memorized physical information (descriptions). Once derived (by the human visual system) from the observable world and learned to be associated with a particular word, these physical information descriptions are soldered in into the knowledgebase. Object recognition, thus, turns out to be a comparison and similarity test between currently acquired physical information and the one already retained in the memory. If the similarity test is successful, starting from this point in the hierarchy and climbing back up on the knowledgebase ladder we will obtain: first, the linguistic label for a recognized object; second, the position of this label (word) in the context of the whole story; and third, the ability to verify the validity of an initial guess by testing the appropriateness of the neighboring parts composing the object, that is, the context of a story. In this way, object's meaningful categorization can be reached, and the first stage of image annotation can be successfully accomplished, providing the basis for farther meaningful (semantic) image interpretation.

One question has remained untouched in our discourse: How does this artificial knowledgebase have to be initially created and brought into our thinking machine disposal? This subject deserves a special discussion.

**4.3 Can Semantic Knowledge be Learned?**

There is no need to reiterate the dictums of the today's Internet revolution, where access and exchange of semantic information is viewed as a prime and an ultimate goal. Machines are supposed to handle the documents' semantic content, and Machine Learning techniques, thus, supporting this knowledge mining venture are supposed to be the leading force, the centre forward of this exciting enterprise. Semantic Knowledge mining is now the hottest topic of every conference discussion, most recent research projects and many other applied science initiatives. However, in the light of our new definition of information, which was recently introduced in my work and re-introduced in the Section 3 of this paper, I am really skeptic about the Machine Learning ability to meet this challenge.

Again, some philosophy would not be out of place here. At first, it must be reiterated that semantics is not a natural property of an image (or a natural property of the data, if you would like a more general view on the subject). Semantics is a property of a human observer that watches and scrutinizes the data, and this property is shared among the observer and other members of his community. By the way, this community does not have to embrace the whole mankind, it can be even a very small community of several people or so, which, nevertheless, were lucky to establish a common view on a particular subject and a common understanding of its meaning. That is the reason why this particular (privet) knowledge can not be attained in any reasonable way, including Machine Learning techniques and tricks.

On the other hand, an intelligent information-processing system has to have at its disposal a memorized knowledgebase hierarchy (implemented, as we postulate, as a collection of typical stories) against which the physical information of its input sensors is matched and associated. Finding the suitable story whose attributes most closely match the sensors' physical information is equivalent to finding the interpretation for the input sensor data (the input physical information). That is the novelty of our proposed architecture. That is the most important feature of our design approach: The knowledgebase hierarchy is used for a linguistic input interpretation, but this knowledge is not derived (by the system) from the input data. It is not learned from the data. On the contrary, in accordance with the top-down information unfolding principle, the knowledge-base hierarchy (as a whole) has to be transferred to the system disposal from the outside. The system doesn't learn the knowledgebase, it is taught to use the knowledgebase (In our case, a pool of task related stories and their ramifications putted at system disposal in advance).

Thus, providing the system with the needed new knowledge each time when the system is due for a new task accomplishment is becoming a natural duty of Artificial Intelligence (Machine Intelligence) system designer. This shift from Machine Learning to Machine Teaching paradigm should become the key point of intelligent system design and development roadmap. But unfortunately, that has not happen although it has been about three years old since the idea was at first articulated and even occasionally published (Diamant, 2006b).

**4.4 Some additional remarks**

That is a very important and an interesting twist in the philosophy of intelligent artificial systems design. It does not result from the understanding of the principals of biological systems intelligence or other proudly declared biological inspirations. On the contrary, it results from pure mathematical considerations of the Kolmogorov's complexity theory. Only now, drawing on the insights of Kolmogorov's theory, we can seize the interpretation of the facts that we usually come across in our natural (biological) surrounding.

It is a very subtle issue among the topics of machine intelligence that I would like to address. "Biologically inspired" means that we borrow from the nature some fruitful ideas, which we intend to replicate in our artificial designs. That is, we presume that we understand or at least are very close to the state of understanding how some biological mechanisms operate, performing their natural duties. But that is not true!.. We don't have even a slightest hint about how the nature works. What we have are gambling guesses, intuitive feelings, speculations, and – nothing more than that.

Another important remark in this regard, is that Nature is not an Engineer. It does not invent new mechanisms and new solutions for its problem-solving. On the contrary, it gradually adjusts and adapts what it already has on the hand. Although the final results are really remarkable, it takes a lot of time to reach them in the course of natural evolution, millions and billions of years. Despite all this, the nature has never reached some very important human-life-shaping revelations – for example, the wheel (as a means for transportation), the cooked food, the writing and numbering practice, etc.

The inventors of "Genetic Programming" provide very interesting quotations from Turing's early works considering Machine Intelligence (Koza et al., 1999; Koza et al., 2002). In his 1948 essay "Intelligent Machines" Alan Turing has identified three broad approaches by which machine intelligence could be achieved: "One approach… is a search through the space of integers representing candidate computer programs, (a logic-driven search)… Another approach is the "cultural search" which relies on knowledge and expertise acquired over a period of years from others. This approach is akin to present-day knowledge-based systems… The third approach is "genetical or evolutionary search"…" (Koza, et al., 1999). From the three, the inventors of Genetic Programming pick up the idea of biological relevance to the problem of machine intelligence acquisition. However, from our point of view (from the point of view inspired by Kolmogorov's theory) this can not be true. Genetic Programming and Neural Networking are pure bottom-up information-processing approaches. As we know today, the right way of information retrieval is a top-down coarse-to-fine approach. Therefore, it might be more intelligent to embrace the first Turing's alternative – the logic-driven approach. That is, relying on pure logical and engineering approaches to find out the best ways of intelligence reification, and only then to verify our hypothetical solutions against known (or unknown) biological evidences and facts. That is exactly what we are intended to do now.

The first issue is the bottom-up versus top-down information-processing alternatives. Despite the traditional dominance of the bottom-up approach, evidence for top-down preliminary processing in biological vision systems is present in research literature since the early 80s of the previous century (Navon, 1977; Chen, 1982). Unfortunately, they were overlooked both by biological and computer vision communities.

The next phenomenon which is usually misunderstood (and consequently mistreated) is the knowledge transfer (in human and animal world), which is usually mistakenly defined as a Learning process. We have proposed a more suitable definition – a Teaching process. Indeed, it turns out that in nature, teaching is a universal and a wide-spread phenomenon. Only recently this fact has become recognized and earned its careful investigation (Csibra, 2007; Hoppitt et al., 2008). Teaching in nature does not mean human-like mentoring – animals do not possess spoken language capabilities. Teaching in nature assumes specific semantic knowledge transfer, specific information relocation from a teacher to a pupil, from one community member to another. And examples of this knowledge transfer are really abundant in our surrounding, if only we are ready to look at them and see them in a proper way.

In this regard, dancing bees that convey to the rest of the hive the information about melliferous sites (Zhang et al., 2005), ants that learn in tandem (Franks & Richardson, 2006), and even bacteria developing their antibiotic resistance as a result of a so-called horizontal gene transfer when a single DNA fragment of one bacteria is disseminated among other colony members (Lawrence & Hendrickson, 2003), all these examples convincingly support our claim that meaningful information (the semantic knowledge base) is always transferred to the individual information processing system from the outside, from the external world. The system does not learn it in a traditionally assumed Machine Learning manner.

Another benefit which biological science can gain from our logically-driven (engineering) approach is the issue of astrocyte-neuron communication. Only defining information as a description message you can explain how astrocities, (the dominant glial cells), "listen and talk" with neuronal and synaptic networks. In their paper, Voltera & Meldolesi wrote that: "One reason that the active properties of astrocytes have remained in the dark for so long relates to the differences between the excitation mechanisms of these cells and those of neurons. Until recently, the electrical language of neurons was thought to be the only form of excitation in the brain. Astrocytes do not generate action potentials, they were considered to be non-excitable and, therefore, unable to communicate. The finding that astrocytes can be excited non-electrically has expanded our knowledge of the complexity of brain communication to an integrated network of both synaptic and non-synaptic routs" (Voltera & Meldolesi, 2005). That is, traditional belief that a spiking neuron burst is a valid form of information exchange and representation does not hold any more, and has to be replaced by a chemical molecular-language-based discription-massages transfer mechanism.

A very important issue of our discussion about semantic information processing is the issue of knowledge representation. As it was already mentioned above, and it also stems from the insights of Kolmogorov's theory, the best form of knowledge representation has to be a language-based description, a narrative, a story. I do not intend to expand here on the implementaition deatails of this issue. I would like to continue to maintain our discussion on a philosophical level. What follows from this is that we have always to consider intelligence as being carried out in a language, in a linguistic structure. That is, although the block-schema depicted in Fig. 2 outlines only visual information incorporation into the semantic processing hierarchy, you can easily imagin physical information of other modalities (hearing, haptics, olfactory senses information) being subjected (usually in parallel with information from other sensors) as attributes of semantic (linguistic) objects into the knowledgebase processing hierarchy. (That will again explain you why functional Magnetic Resonance Imaging shows you that visual stimuli are processed in audio stimuli processing zones, which are naturally associated with speech processing. The simple explanation for this is that all modalities are finally processed in the language processing zone, as it is proposed by our approach.)

The next important issue, which naturally follows the preceeding ones, is the narrative story form of knowledge representation accepted for the discussed case of semantic information processing. Linguistic tagging, that means labeling image objects with words, is a well known and widely used methodology of image semantics retrival supported by a special class of Machine Learning techniques (Barnard et al., 2003; Duygulu et al., 2008; Blondin Masse et al., 2008). This way of thinking naturally stems from another wide-spread assumption that ontology (the basis of semantic reasoning and elaboration) is a vocabulary, a thesaurus, a dictionary. What follows from our descriptive form of knowledge representation is that ontology has to be treated as a story, a narrative, which naturally describes the system's behavior in various real-life-encountered situations. However, this very important aspect of intelligent systems design philosophy leads us far away from the main theme of our discussion – the philosophy of Machine Learning. And for that reason I will quit at this point, and not continue further.

## 5. Conclusions

In this paper I have attempted to promote a new Thinking Machines design and development philosophy. The central point of my approach is a new definition of information, that is, a notion of information as a language-based description. Then, above it the notion of intelligence can be placed, defining intelligence as the system's ability to process information. The principles of information mining should be placed in the lower part of the construction. Thus, it seems to me, a proper frame for a rational Artificial or Machine Intelligence devices research and development enterprise can be established.

Essentially, the declared focus of the paper's subject is the issue of Machine Learning, which is assumed to be a bundle of techniques used to support all information-processing machinery. But, as you know, Machine Learning as by now (and already for a very long time) is treated as an independent and stand alone discipline, totally detached from its original destination – Thinking Machines research and development (Turing, 1950). The roadmap for this challenge was formulated at the Dartmouth College meeting in the summer of 1956 (McCarthy, et al. 1955). The date of this meeting is considered today as the Artificial Intelligence (AI) birthday. (The very name of AI was coined at this time by John McCarthy, one of the authors of the Dartmouth Proposal).

At first, the excitement and hopes were really high, and the goals have seemed to be reachable in a short while. In the Panel Discussion at the Artificial General Intelligence (AGI) Workshop in 2006, Steve Grand has recalled that "Rodney Brooks has a copy of a memo from Marvin Minsky (another father of the Dartmouth Proposal), in which he suggested charging an undergraduate for a summer project with the task of solving vision. I don't know where that undergraduate is now, but I guess he hasn't finished yet" (Panel Discussion, 2006).

Indeed, problems of Vision, as well as all other AI troubles, have turned out to be much more complicated and problematic than it looked out at the beginning. Within a decade or so, it became clear that AI tribulations are immense, maybe even intractable. As a consequence, the AI community to a large extent has abandoned its original dream, and turned to more "practical" and "manageable" problems (Wang & Goertzel, 2006). "AI has evolved to being a label on a family of relatively disconnected efforts" (Brachman, 2005).

Exactly the same were the milestones of Machine Learning. Machine Learning, which was always perceived as an indispensible component of intelligence, has undergone all the metamorphoses as its general domain. At first, there was a generous offer to let the system by itself (in an autonomous manner) to find out the best way to mimic Intelligence. Why to trouble oneself trying to grasp the principles of intelligence? Let us give the machine the chance to do this by itself. (I can not to withstand the temptation to provide an example of such a fatal misunderstanding: IGI Global Publisher (formerly Idea Group Inc.) has published a Call for Chapter Proposals for a future book "Intelligent Systems for Machine Olfaction: Tools and Methodologies" (Can be found at the publisher site: http://www.igi-global.com/requests/details.asp?ID=610). You can read in the Introduction part of it: "Intelligent systems are those that, given some data, are able to learn from that data. This ability makes it possible for complex systems to be modeled and/or for performance to be predicted. In turn it is possible to control their functionality through learning/training, without the need for a priory knowledge of the system's structure". Once more, I apologize for such a so long quotation.)

Then, when the first idealistic objective has failed, Machine Learning was broken into pieces, disintegrated and fragmented to many partial tasks and goals. Now the question in the paper's title – "When and Where the Horses Went Astray?" – can be answered beyond any doubts: It has happened about 50 years ago!

From the standpoint that we possess today, we can even spell out the fundamental flaws which are responsible for this derailment: First, the bottom-up philosophy of information retrieval. (As we know today, the right way of information treatment is the top-down coarse-to-fine line of information processing). Second, is the lack of a proper definition of information, leading, consequently, to a lack of a clear distinction between physical and semantic information. (This failure had a tremendous impact on the Machine Learning disruption). The same can be said about the third misleading factor – misunderstanding of the very nature of semantic information, which has led to an endless, infamous race for knowledge and semantic meaning extraction directly from the raw data. (Which is, obviously, a philosophical lapse).

For the same reasons, the basic notion of intelligence has been overlooked and defined erroneously. I hope, in this paper I was lucky to repair some of these misconceptions.